\documentclass{article}

\usepackage{microtype}
\usepackage{graphicx}
\usepackage{subcaption}
\usepackage{booktabs} %
\usepackage{tabularx}

\usepackage{mlmacros}

\variables{u,x,z}
\probdists{p,q}

\usepackage[T1]{fontenc}
\usepackage[acronym,nomain,
    style=list,
    nonumberlist,
    shortcuts,
    smallcaps,
]{glossaries}

\glsdisablehyper %
\newacronym[firstplural=variational autoencoders, plural=VAEs]{VAE}{vae}{variational autoencoder}
\newacronym{RL}{rl}{reinforcement learning}
\newacronym{GENESIS}{genesis}{\textsc{gene}rative \textsc{s}cene \textsc{i}nference and \textsc{s}ampling}
\newacronym{SGMM}{sgmm}{spatial Gaussian mixture model}
\newacronym{SBD}{sbd}{spatial broadcast decoder}
\newacronym{GECO}{geco}{Generalised \textsc{elbo} with Constrained Optimisation}
\newacronym[firstplural=multilayer perceptrons, plural=MLPs]{MLP}{mlp}{multilayer perceptron}
\glsunset{MLP}
\newacronym[firstplural=recurrent neural networks, plural=RNNs]{RNN}{rnn}{recurrent neural network}
\glsunset{RNN}

\newacronym{BN}{bn}{batch normalisation}

\newacronym{ARI}{ari}{Adjusted Rand Index}
\newacronym{MSC}{msc}{Mean Segmentation Covering}

\usepackage{hyperref}

\usepackage[accepted]{ool2020}

\icmltitlerunning{Reconstruction Bottlenecks in Object-Centric Generative Models}

\begin{document}

\twocolumn[
\icmltitle{Reconstruction Bottlenecks in Object-Centric Generative Models}

\begin{icmlauthorlist}
\icmlauthor{Martin Engelcke}{a2i}
\icmlauthor{Oiwi Parker Jones}{a2i}
\icmlauthor{Ingmar Posner}{a2i}
\end{icmlauthorlist}

\icmlaffiliation{a2i}{Applied AI Lab, University of Oxford, UK}
\icmlcorrespondingauthor{Martin Engelcke}{martin@robots.ox.ac.uk}

\icmlkeywords{Object representations, generative models, representation learning, variational autoencoders}

\vskip 0.3in
]

\printAffiliationsAndNotice{}  %

\begin{abstract}

A range of methods with suitable inductive biases exist to learn interpretable object-centric representations of images without supervision.
However, these are largely restricted to visually simple images; robust object discovery in real-world sensory datasets remains elusive.
To increase the understanding of such inductive biases, we empirically investigate the role of ``reconstruction bottlenecks'' for scene decomposition in \textsc{genesis}, a recent \textsc{vae}-based model.
We show such bottlenecks determine reconstruction and segmentation quality and critically influence model behaviour.

\end{abstract}

\section{Introduction}

\setcounter{footnote}{1}

Interest in unsupervised object-centric generative models \citep{burgess2019monet,greff2019multi,engelcke2020genesis} is driven by the premise of increased sample efficiency and generalisation for tasks that involve interaction with objects \citep[e.g.][]{watters2019cobra}.
While these methods exhibit suitable inductive biases to identify interpretable components, understanding of these biases is limited and their application to more complex real-world datasets is an unresolved challenge \citep[e.g.][]{greff2019multi}.

We posit that enhancing the understanding of inductive biases for scene decomposition can facilitate the development of better object-centric generative models, both on current and more difficult datasets.
Specifically, we argue that methods based on \glspl{VAE} \citep{kingma2013auto,rezende2014stochastic} that learn object-centric representation by learning to reconstruct input images \citep[e.g.][]{burgess2019monet,greff2019multi,engelcke2020genesis} feature what we call \emph{reconstruction bottlenecks} that induce decomposition by prohibiting the model from reconstructing an image as a single component.
Therefore, we present an empirical investigation of this mechanism as the reconstruction bottleneck is traversed and we examine how it impacts reconstruction and segmentation quality.\!\footnotemark

\newpage
\footnotetext{This is distinct from the ``information bottleneck theory'' \citep{shwartz2017opening} on generalisation.}

To do this, we conduct experiments with \acrshort{GENESIS} \citep{engelcke2020genesis}, a recently developed model for unsupervised segmentation and component-wise generation.
\acrshort{GENESIS} features two sets of \glspl{VAE}: \emph{mask \glspl{VAE}} model pixel-wise segmentation masks and \emph{component \glspl{VAE}} reconstruct object appearances.
Notably, the model features an \emph{asymmetric architecture}; the mask \glspl{VAE} feature mirrored convolutional and deconvolutional layers, but the component \glspl{VAE} use \glspl{SBD} \citep{watters2019spatial}.
These encourage latent disentanglement and act---as we argue---as a reconstruction bottleneck.
We thus examine model behaviour when varying the latent dimensions before the object appearance decoder with the original architecture as well as with a modified, \emph{symmetric architecture}.

Experiments are conducted on \emph{Multi-dSprites} \citep{burgess2019monet}, \emph{ShapeStacks} \citep{groth2018shapestacks}, and \emph{ObjectsRoom} \citep{multiobjectdatasets19}.
The original, asymmetric \acrshort{GENESIS} architecture achieves similar segmentation performance across a broad range of latent dimensions and only degrades when it is very small.
We argue the \gls{SBD}-architecture inhibits the learning of good reconstructions and thus serves as the \emph{effective bottleneck} when the latent dimensionality is large.
For the symmetric architecture with matched encoders and decoders, however, good segmentation is only achieved in a very narrow window of suitably small latent dimensions before the object appearance decoder.

An intricate interplay between segmentation and reconstruction is observed: If the bottleneck is too narrow, segmentation and reconstruction degrade.
If it is too wide, reconstruction is reasonable but the model collapses to a single component and no useful segmentation is learned.
While these observations are intuitive for experienced practitioners, we found that---to the best of our knowledge---an extensive formal treatment has been missing in the literature.
They also provide insight into the suitability of \glspl{SBD} for object-centric models trained on these types of datasets.
We believe this provides useful guidance for researchers and practitioners, both established and new to the field, and hope this work encourages a more open discussion of the design process of object-centric models.\!\footnote{Experiments can be reproduced with the code \href{https://github.com/applied-ai-lab/genesis}{here}.}

\newpage
\section{Related Work}

Object-centric generative models can be categorised according to their training objective: (1) reconstruction-based \citep{huang2015efficient,eslami2016attend,greff2016tagger,greff2017neural,greff2019multi,van2018relational,kosiorek2018sqair,kosiorek2019stacked,crawford2019spatially,burgess2019monet,von2020towards,kossen2020structured,lin2020space,jiang2020scalor,engelcke2020genesis} and (2) adversarial \citep{van2018case,chen2019unsupervised,bielski2019emergence,arandjelovic2019object,azadicompositional,nguyen2020blockgan}.
A discriminative approach for learning object-centric representations has also been proposed \citep{kipf2019contrastive}.

While this work focuses on the reconstruction-based model in \citet{engelcke2020genesis}, we argue most methods of this type feature reconstruction bottlenecks for inducing scene decomposition.
For example, \citet{burgess2019monet} also use a flexible segmentation network and a less expressive component appearance network with an \gls{SBD}.
While \citet{greff2019multi} use the same network for segmentation and appearance, they also use an \gls{SBD} with limited expressiveness for decoding.
Similarly, in methods that use spatial transformers \citep{jaderberg2015spatial} instead of segmentation masks for distinguishing objects \citep[e.g.][]{huang2015efficient,eslami2016attend,kosiorek2018sqair,crawford2019spatially,jiang2020scalor}, the dimensions of the transformer sampling grid constitute a reconstruction bottleneck.
We therefore conjecture the findings in this work will also apply to a range of other model formulations.

\section{Background: GENESIS}

\gls{GENESIS} \citep{engelcke2020genesis} is a \gls{VAE} which encodes an image $\bx \in \RR^{H \times W \times C}$ into a set of \emph{mask variables} $\bz^m_k$ and a set of \emph{component variables} $\bz^c_k$ where $k \in 1,\dots,K$ and $K$ is the maximum number of image components.
The mask variables are decoded into a set of normalised \emph{mixture probabilities}, or segmentation masks, $\pi_k$ and the component variables are decoded into a parameterised set of generated component appearances \p{\bx_k}{\bz^c_k}{\theta}.
Together, these parameterise a \gls{SGMM} for the image likelihood \citep[see][]{greff2017neural}:
\begin{equation}
    \p{\bx}{\bz^m_{1:K}, \bz^c_{1:K}} = \sum_{k=1}^K \pi_k\, \p{\bx_k}{\bz^c_k}{\theta}\,.
\end{equation}

\begin{figure}[t]
    \centering
    \includegraphics[trim=0 0 45 0, clip, width=\linewidth]{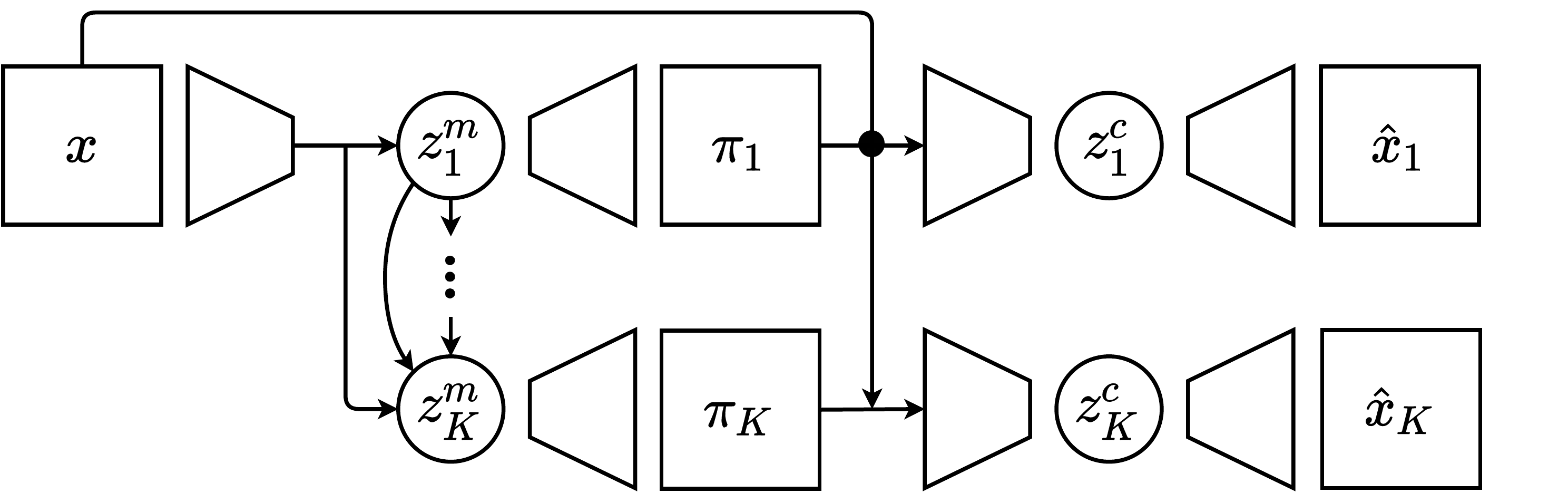}
    \caption{\gls{GENESIS} architecture \citep{engelcke2020genesis}. Given an image $\bx$, mask variables $\bz^m_k$ and component variables $\bz^c_k$ encode mixture probabilities $\pi_k$ and component appearances \p{\bx_k}{\bz^c_k}{\theta}.}
    \label{fig:genesis_diagram}
\end{figure}

The mask variables are inferred in an autoregressive fashion and decoded in parallel.
Individual masks are subsequently concatenated with the image to infer the component latents.
This forward pass through the model is illustrated in Fig. \ref{fig:genesis_diagram}.

\textbf{Generative model}\ \
The generative model can be written as:
\begin{equation}
    \p{\bx}{}{\theta} = \iint \p{\bx}{\bz^c,\bz^m}{\theta} \p{\bz^c}{\bz^m}{\theta} \p{\bz^m}{}{\theta} \dint \bz^m \dint \bz^c\,,
\end{equation}
where component subscripts are not shown for brevity.
The prior distributions over latent variables factorise as:
\begin{equation}
    \p{\bz_{1:K}^m}{}{\theta} = \prod_{k=1}^K \p{\bz^m_k}{\bz^m_{1:k-1}}{\theta}\,,  \label{eq:mask_prior}
\end{equation}
\begin{equation}
    \p{\bz^c_{1:K}}{\bz^m_{1:K}}{\theta} = \prod_{k=1}^K \p{\bz^c_k}{\bz^m_k}{\theta}\,. \label{eq:appearance_prior}
\end{equation}

\textbf{Inference model}\ \
The inference model can be written as:
\begin{equation}
	\q{\bz_{1:K}^c, \bz_{1:K}^m}{\bx}{\phi} = \q{\bz_{1:K}^m}{\bx}{\phi}\, \q{\bz_{1:K}^c}{\bx, \bz_{1:K}^m}{\phi}\,,
\end{equation}
The posterior distributions over latent variables factorise as:
\begin{equation}
	\q{\bz_{1:K}^m}{\bx}{\phi} = \prod_{k=1}^K \q{\bz_k^m}{\bx, \bz_{1:k-1}^m}{\phi}\,,  \label{eq:mask_posterior}
\end{equation}
\begin{equation}
	\q{\bz_{1:K}^c}{\bx, \bz_{1:K}^m}{\phi} = \prod_{k=1}^K \q{\bz_k^c}{\bx, \bz_{1:k}^m}{\phi}\,. \label{eq:appearance_posterior}
\end{equation}

\textbf{Implementation}\ \
All prior and conditional distributions are diagonal Gaussians.
The distributions in Eq. \ref{eq:appearance_prior} and Eq. \ref{eq:appearance_posterior} are parameterised by feedforward networks and the autoregressive distributions in Eq. \ref{eq:mask_prior} and Eq. \ref{eq:mask_posterior} are parameterised by \glspl{RNN}.
The mask sub-network uses gated (de-)convolutions \citep{dauphin2017language,berg2018sylvester} with \gls{BN} \citep{ioffe2015batch} and the component sub-network has an architecture similar to \citet{burgess2019monet} with an \gls{SBD} and ELUs \citep{clevert2015fast} instead of ReLUs \citep{glorot2011deep}.
This \emph{architecture asymmetry} is at the focus of the experiments in this work.
Finally, the model is trained by optimising the \gls{GECO} objective from \citet{rezende2018taming}.
We refer the reader to \citet{engelcke2020genesis} for further details.

\section{Experiments}

\begin{figure*}[t!]
    \centering
    \begin{subfigure}{.33\textwidth}
        \includegraphics[width=\textwidth]{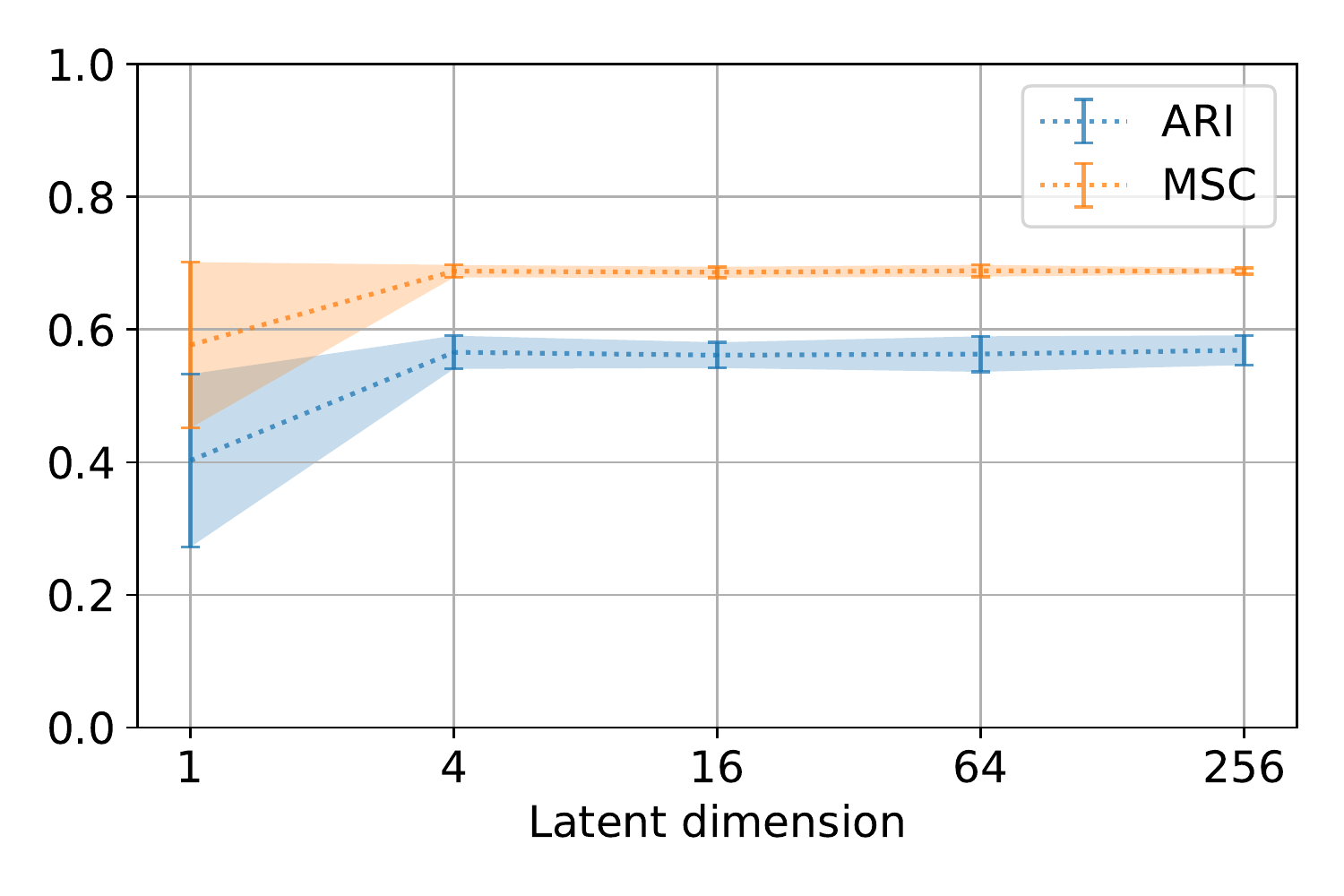}
        \caption{Multi-dSprites}
    \end{subfigure}
        \begin{subfigure}{.33\textwidth}
        \includegraphics[width=\textwidth]{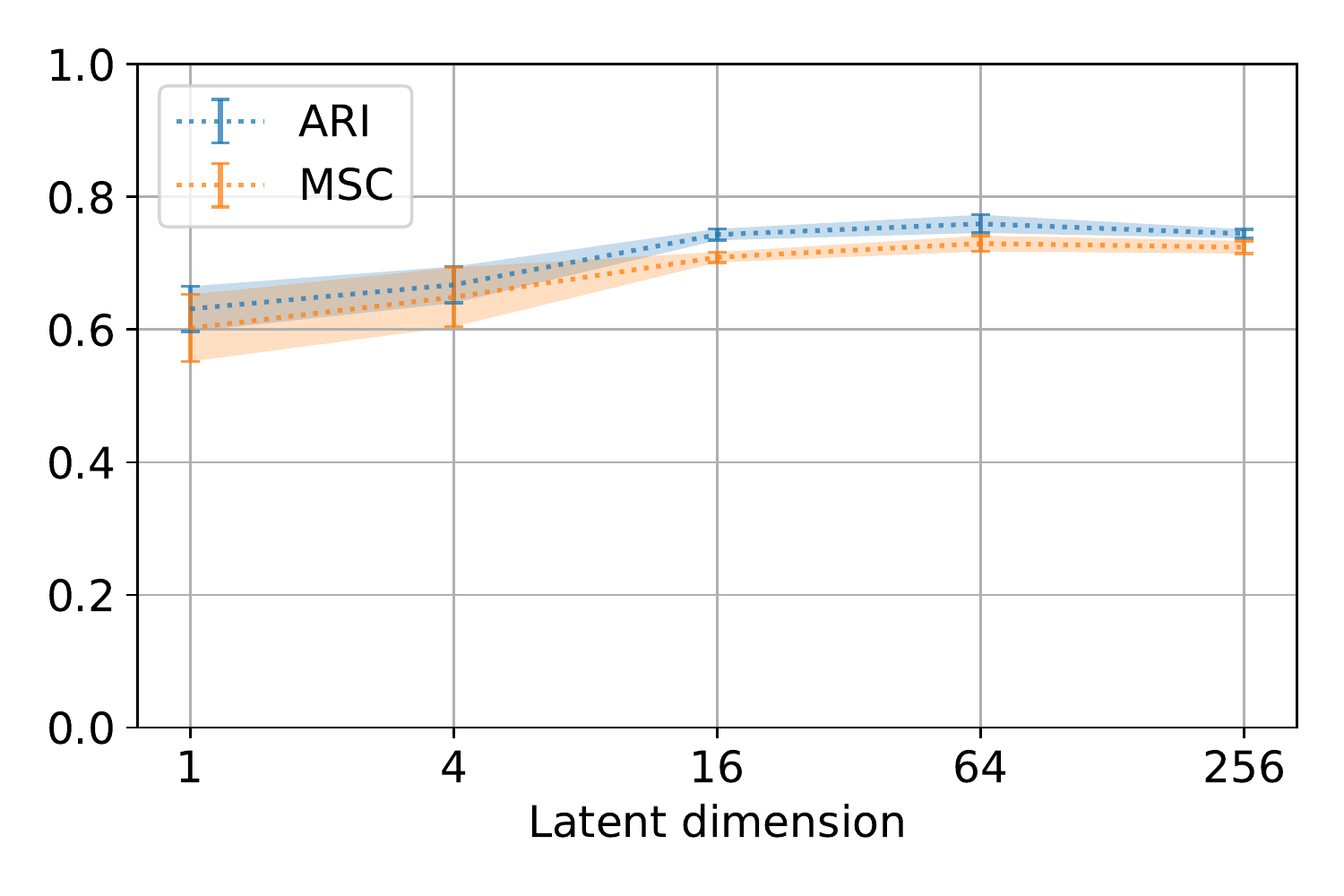}
        \caption{ShapeStacks}
    \end{subfigure}
    \begin{subfigure}{.33\textwidth}
        \includegraphics[width=\textwidth]{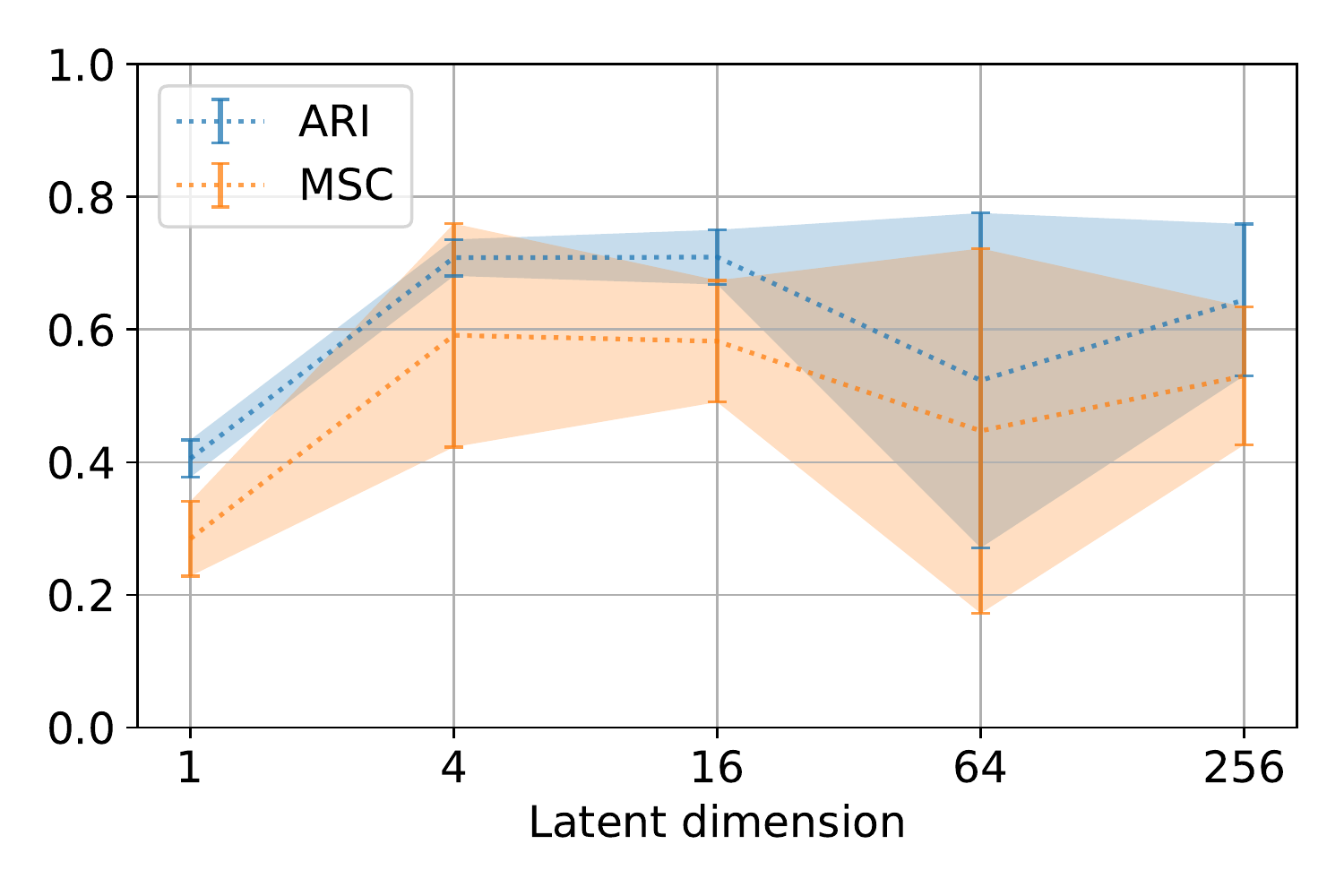}
        \caption{ObjectsRoom}
    \end{subfigure}
    \caption{Segmentation performance for the original---asymmetric---\gls{GENESIS} architecture as the component latent dimension is varied. Decomposition occurs throughout the entire range for all three datasets, but performance drops when the latent dimension is very small.}
    \label{fig:quantitative_asymmetric}
\end{figure*}

\textbf{Datasets}\ \
Experiments are conducted on Multi-dSprites, \citep{burgess2019monet}, ShapeStacks \citep{groth2018shapestacks}, and ObjectsRoom \citep{multiobjectdatasets19}.
A withheld validation set is used for obtaining segmentation metrics.
We follow \citet{engelcke2020genesis} for Multi-dSprites and ShapeStacks.
For ObjectsRoom, we withhold 20,000 images from the training set in \citet{multiobjectdatasets19} for evaluation.

\textbf{Setup}\ \
We perform experiments with different latent dimensions in the component \gls{VAE} for both the original, asymmetric architecture from \citet{engelcke2020genesis} (Sec. \ref{sec:asymmetric}) and a modified, symmetric architecture (Sec. \ref{sec:symmetric}).
The asymmetric architecture has a spatial broadcast decoder (\gls{SBD}) in the component \gls{VAE} and gated (de-)convolutions and batch normalisation in the mask \gls{VAE}; experiments are conducted with latent dimensions of $\{1, 4, 16, 64, 256\}$.
For the symmetric architecture, the same encoder and decoder with gated \mbox{(de-)convolutions} and batch normalisation as used in the mask \gls{VAE} are also used in the component \gls{VAE}.
Here, we observed that decomposition is only achieved for smaller latent dimensions and therefore modified the range to $\{1, 2, 4, 8, 16\}$ to increase the resolution in the hyperparameter subspace where interesting behaviour occurs.
Experiments use the implementation from \citet{engelcke2020genesis} and identical optimisation hyperparameters.

\textbf{Metrics}\ \
Segmentation performance is quantified with the \gls{ARI} and the \gls{MSC}, on 300 images not seen during training \citep[see][]{greff2019multi,engelcke2020genesis}.
We report the means and standard deviations obtained with three different random seeds.
Larger values are better.

\textbf{Compute}\ \
Training a single model takes on average around two days on an NVIDIA Titan RTX card.
The results presented in this work amount to training a total of 90 individual models, corresponding to ca. 180 GPU days.

Further details and results are included in the appendix.

\subsection{Asymmetric architecture}
\label{sec:asymmetric}

Segmentation performance on the three datasets is shown in Fig. \ref{fig:quantitative_asymmetric}.
It can be observed that the models decompose the images across the whole range of latent dimensions.
Separately, we also train the component reconstruction architecture as a vanilla \textsc{sbd-vae} for different latent dimensions, recording either the number of iterations required to reach the \gls{GECO} reconstruction goal or the final moving average of the reconstruction error (Tab. \ref{tab:bd}).
The \textsc{sbd-vae} never reaches the \gls{GECO} goal on its own and the reconstruction error does not further decrease for latent dimensions larger than 32.
We conclude that the structure of the \gls{SBD} slows down the minimisation of the reconstruction loss and that in the asymmetric \gls{GENESIS} architecture it is this structure---rather than latent dimensionality---which acts as the effective bottleneck, at least when the latent dimension is large.
This therefore prevents the models to collapse to a trivial solution where images are reconstructed as a single component.
Finally, segmentation performance slightly drops when the latent dimension is very small.
We conjecture that this is caused by the lack of detail in lower quality reconstructions.

\begin{table}[h!]
    \centering
    \caption{\textsc{sbd-vae} training iterations needed to reach the \gls{GECO} reconstruction goal of 0.5655. If the goal is not reached within 500k iterations, the final moving average error is reported instead.}
    \label{tab:bd}
    \newcolumntype{Y}{>{\centering\arraybackslash}X}
    \begin{tabularx}{1.0\columnwidth}{c c*{6}{Y}}
        \toprule
        & \multicolumn{2}{c}{Multi-dSprites} & \multicolumn{2}{c}{ShapeStacks} & \multicolumn{2}{c}{ObjectsRoom} \\
        \cmidrule(lr){2-3} \cmidrule(lr){4-5} \cmidrule(lr){6-7}
        Latent & Goal & 500k & Goal & 500k & Goal & 500k \\
        dim    & iter & err  & iter & err  & iter & err \\
        1   & - & 0.5885 & - & 0.5852 & - & 0.6064 \\
        2   & - & 0.5820 & - & 0.5819 & - & 0.5921 \\
        4   & - & 0.5763 & - & 0.5785 & - & 0.5820 \\
        8   & - & 0.5730 & - & 0.5754 & - & 0.5767 \\
        16  & - & 0.5697 & - & 0.5730 & - & 0.5715 \\
        32  & - & 0.5676 & - & 0.5709 & - & 0.5703 \\
        64  & - & 0.5675 & - & 0.5712 & - & 0.5698 \\
        128 & - & 0.5676 & - & 0.5715 & - & 0.5700 \\
        256 & - & 0.5674 & - & 0.5710 & - & 0.5700 \\
        \bottomrule
    \end{tabularx}
\end{table}

\begin{figure*}[h!]
    \centering
    \begin{subfigure}{.33\textwidth}
        \includegraphics[width=\textwidth]{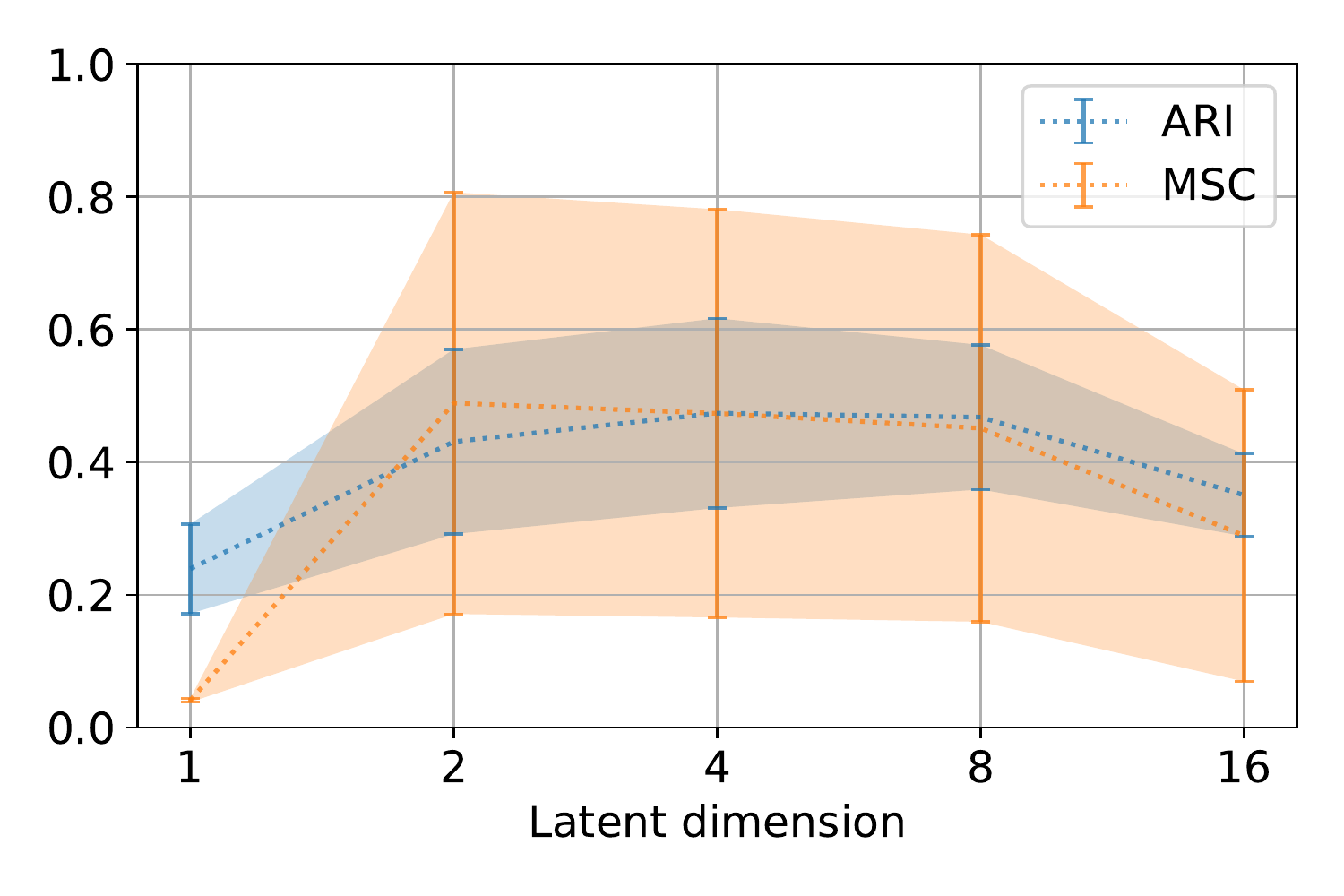}
        \caption{Multi-dSprites}
    \end{subfigure}
        \begin{subfigure}{.33\textwidth}
        \includegraphics[width=\textwidth]{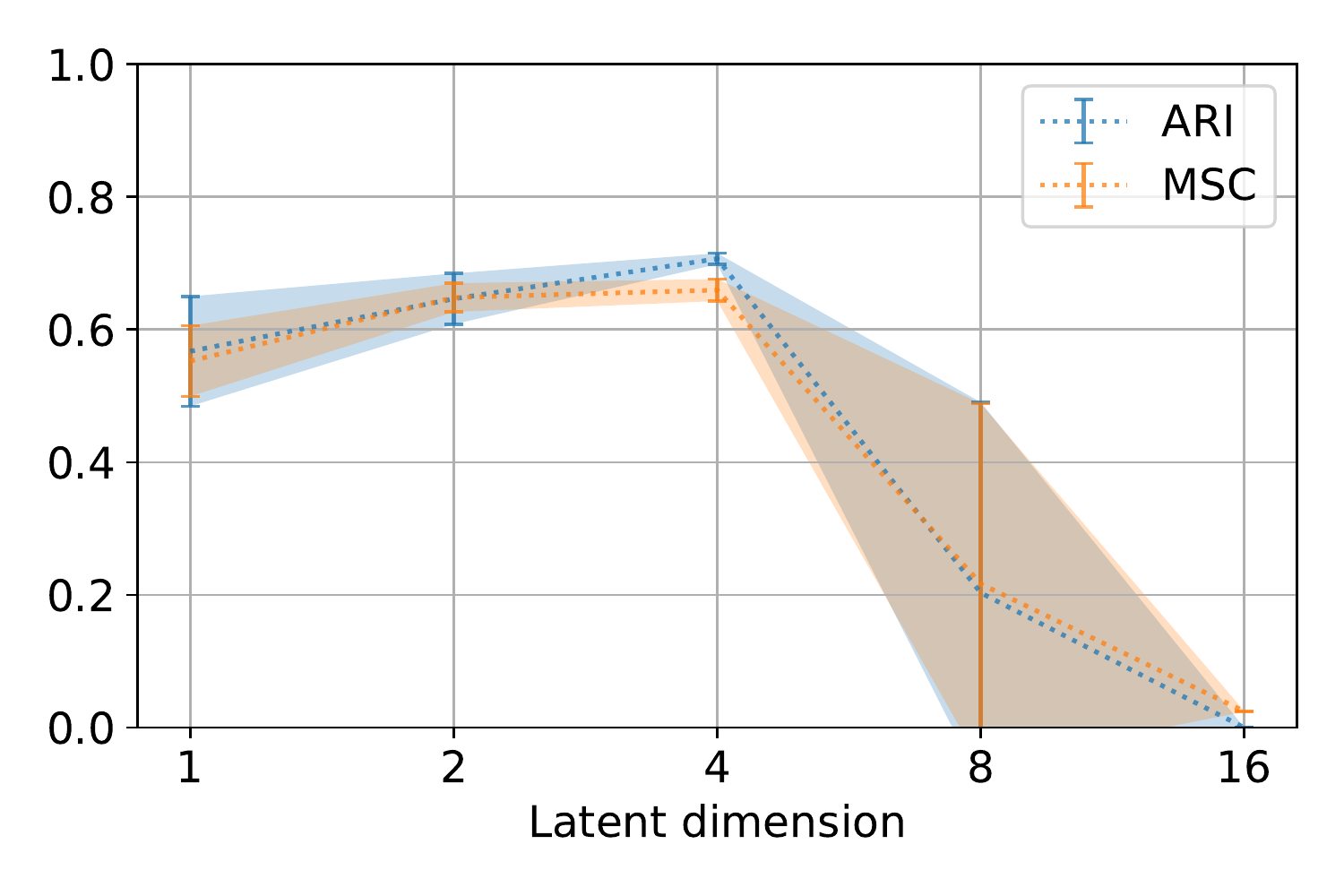}
        \caption{ShapeStacks}
    \end{subfigure}
    \begin{subfigure}{.33\textwidth}
        \includegraphics[width=\textwidth]{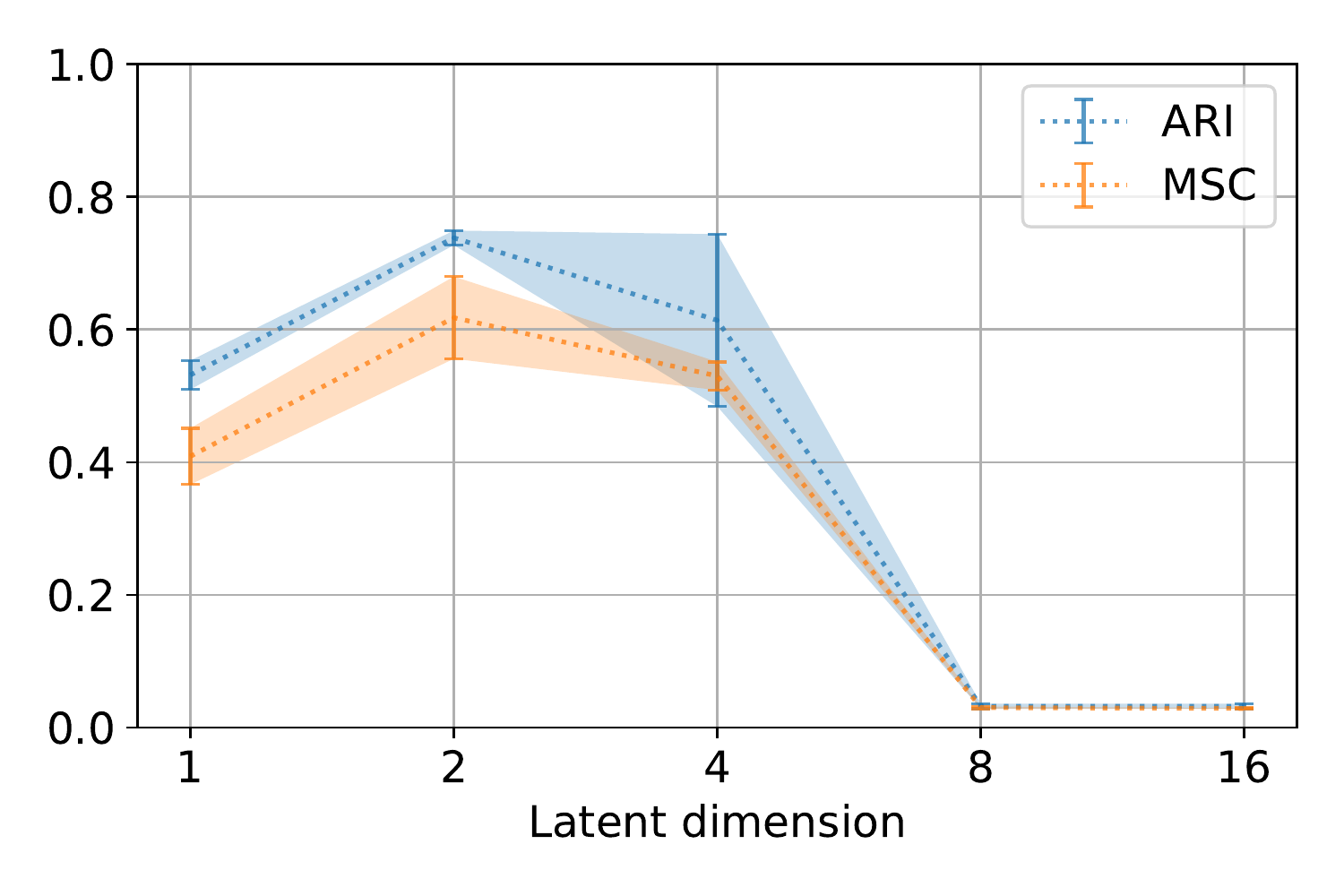}
        \caption{ObjectsRoom}
    \end{subfigure}
    \caption{Segmentation performance for the modified---symmetric---\gls{GENESIS} architecture as the component latent dimension is varied. Decomposition only occurs in a small range of latent dimensions and performance drops when the latent dimension is small.}
    \label{fig:quantitative_symmetric}
\end{figure*}

\subsection{Symmetric architecture}
\label{sec:symmetric}

Segmentation performance is shown in Fig. \ref{fig:quantitative_symmetric}.
It can be seen that if the component latent dimension is too large, the models consistently collapse to a single component.
As shown in the appendix, this does not necessarily adversely affect reconstruction quality: the modified component \gls{VAE} is expressive enough to produce good reconstructions without decomposition.
For smaller latents, however, there is a range where the models learn to segment the images into object-like components.
Similar to before, segmentation quality deteriorates when the latent dimension becomes very small.
We argue that the difference in behaviour here is due to the modified architecture being sufficiently expressive so that the latent dimension is the effective reconstruction bottleneck, hence exhibiting a critical influence on model behaviour.
We train the component architecture as a vanilla \textsc{dc-vae} for different latent dimensions (Tab. \ref{tab:dc}).
The range of latents where \gls{GENESIS} learns to decompose images somewhat correlates with the range where the final \textsc{dc-vae} reconstruction error is relatively far away from the \gls{GECO} goal, i.e. $>0.5670$ for latent dimensions in the interval $[1, 8]$.

\begin{table}[h!]
    \centering
    \caption{\textsc{dc-vae} training iterations needed to reach the \gls{GECO} reconstruction goal of 0.5655. If the goal is not reached within 500k iterations, the final moving average error is reported instead.}
    \label{tab:dc}
    \newcolumntype{Y}{>{\centering\arraybackslash}X}
    \begin{tabularx}{1.0\columnwidth}{c c*{6}{Y}}
        \toprule
        & \multicolumn{2}{c}{Multi-dSprites} & \multicolumn{2}{c}{ShapeStacks} & \multicolumn{2}{c}{ObjectsRoom} \\
        \cmidrule(lr){2-3} \cmidrule(lr){4-5} \cmidrule(lr){6-7}
        Latent & Goal & 500k & Goal & 500k & Goal & 500k \\
        dim    & iter & err  & iter & err  & iter & err \\
        1   & -   & 0.6008 & -    & 0.5820 & - & 0.6082 \\
        2   & -   & 0.5808 & -    & 0.5766 & - & 0.5904 \\
        4   & -   & 0.5753 & -    & 0.5726 & - & 0.5825 \\
        8   & -   & 0.5697 & -    & 0.5683 & - & 0.5733 \\
        16  & -   & 0.5660 & 246k & -      & - & 0.5668 \\
        32  & \phantom{0}79k & - & \phantom{0}25k & - & \phantom{0}61k & - \\
        64  & \phantom{0}21k & - & \phantom{00}9k & - & \phantom{0}16k & - \\
        128 & \phantom{0}10k & - & \phantom{00}6k & - & \phantom{00}7k & - \\
        256 & \phantom{00}8k & - & \phantom{00}5k & - & \phantom{00}6k & - \\
        \bottomrule
    \end{tabularx}
\end{table}
\section{Discussion}

This work performs an empirical investigation into the role of ``reconstruction bottlenecks'' in a \gls{VAE}-based unsupervised object-centric generative model.
An intricate interplay between reconstruction and image decomposition occurs: if the effective bottleneck in the part of the model that reconstructs individual components is too narrow, reconstruction and segmentation quality suffer.
Conversely, if the effective bottleneck is too wide, the models do not learn to decompose images and instead collapse to a trivial solution.
Depending on design decisions made by the practitioner, this effective bottleneck can either be controlled by the latent dimensionality or the architecture.

We believe these results provide valuable insights into the inductive biases for these types of models and provide useful guidance for researchers and practitioners.
It would be interesting to examine these properties with other spatial mixture models \citep[e.g.][]{burgess2019monet,greff2019multi} as well as spatial transformer models \citep[e.g.][]{eslami2016attend,jiang2020scalor}.
Another possible direction could be to investigate the relationship between optimal bottleneck size and dataset complexity as well as utilising vanilla \gls{VAE} behaviour to predict decomposition quality when the identical \gls{VAE} is used for modelling components in an object-centric model.

Reconstruction bottlenecks appear to be a sufficient mechanism for learning object-centric representations when scene components have comparable visual complexity.
It would be interesting to investigate how this approach fares for datasets where texture variance is more variable across scene components; for example, in a real-world image of an indoor room, the walls are often uniformly coloured whereas other components, e.g. furniture, can have arbitrarily complex appearances.
Leveraging additional inductive biases such as motion \citep{kosiorek2018sqair,jiang2020scalor} or depth and geometry \citep{byravan2017se3,nguyen2020blockgan} might be beneficial for this.

\clearpage

\section*{Acknowledgements}
This research was supported by an EPSRC Programme Grant (EP/M019918/1) and an Amazon Research Award.
The authors would like to acknowledge the use of the University of Oxford Advanced Research Computing (ARC) facility in carrying out this work, \url{http://dx.doi.org/10.5281/zenodo.22558}, and the use of Hartree Centre resources.
The authors would also like to thank the reviewers for the thoughtful feedback.

\bibliography{references}
\bibliographystyle{ool2020}

\clearpage
\appendix
\onecolumn
\section{Component \textsc{vae} architecture details}

\subsection{Asymmetric \textsc{genesis} and \textsc{sbd-vae}}

\begin{figure*}[h!]
    \centering
    \begin{subfigure}{.33\textwidth}
        \includegraphics[trim=0 0 0 0, clip, width=\textwidth]{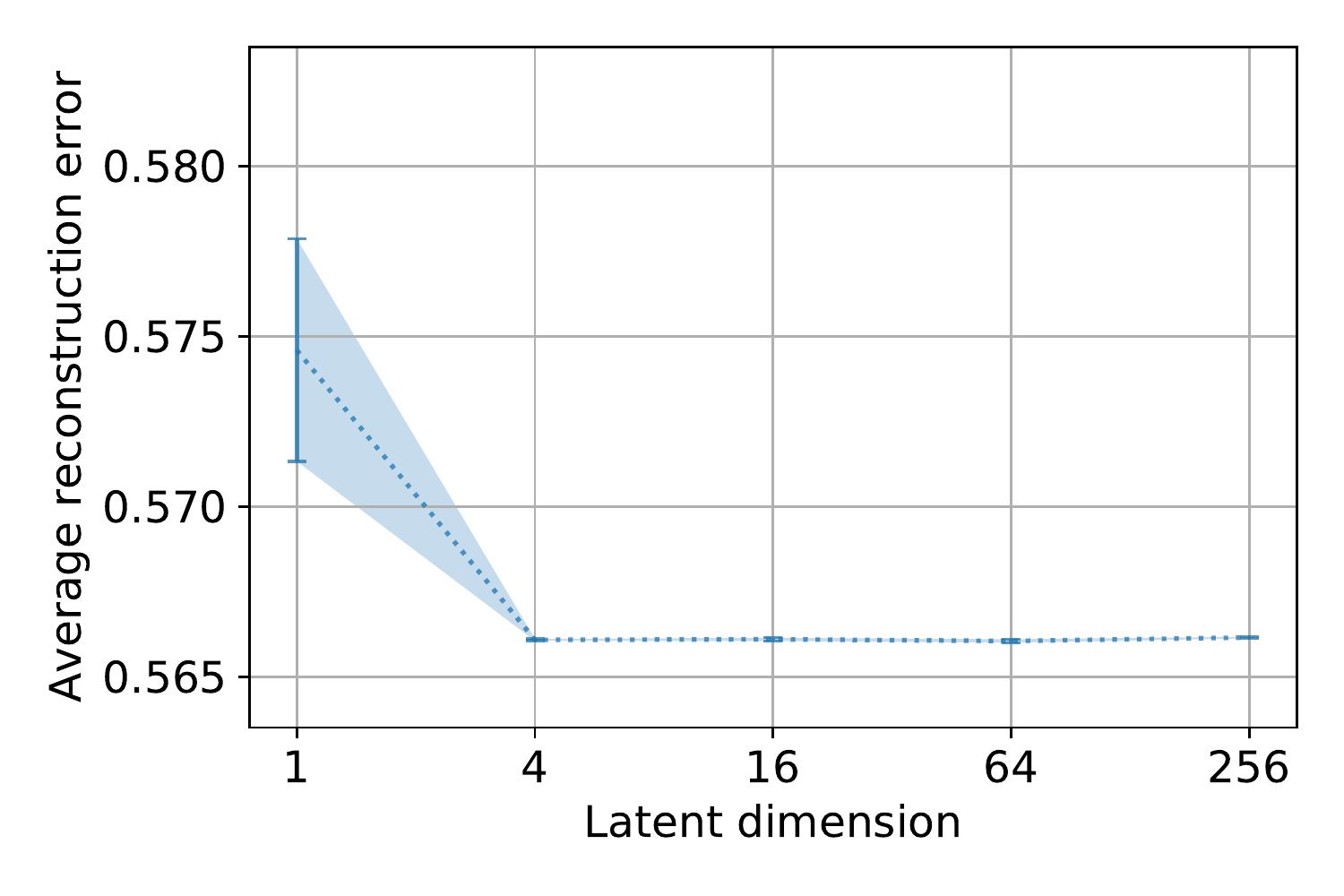}
        \caption{Multi-dSprites}
    \end{subfigure}
        \begin{subfigure}{.33\textwidth}
        \includegraphics[trim=0 0 0 0, clip, width=\textwidth]{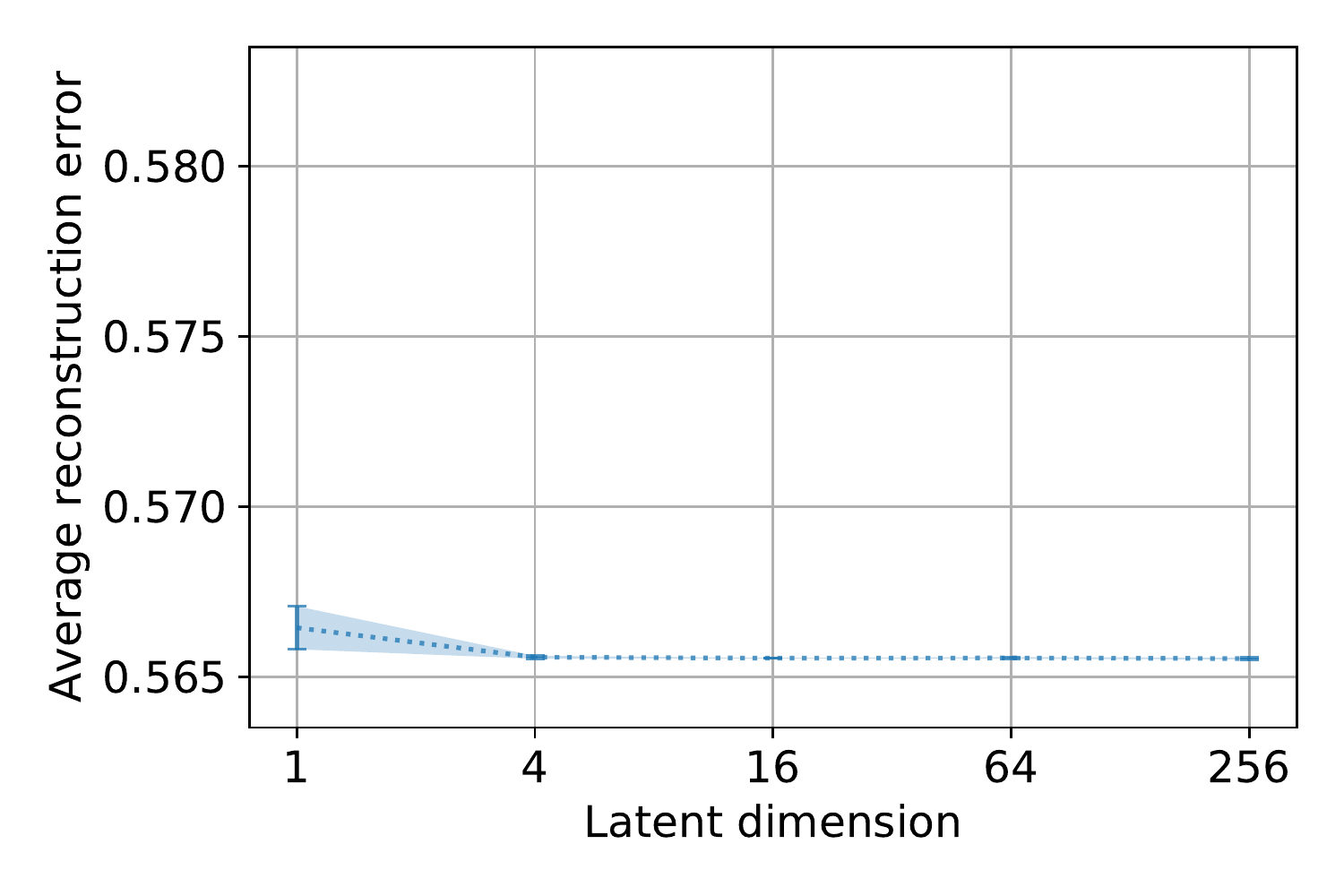}
        \caption{ShapeStacks}
    \end{subfigure}
    \begin{subfigure}{.33\textwidth}
        \includegraphics[trim=0 0 0 0, clip, width=\textwidth]{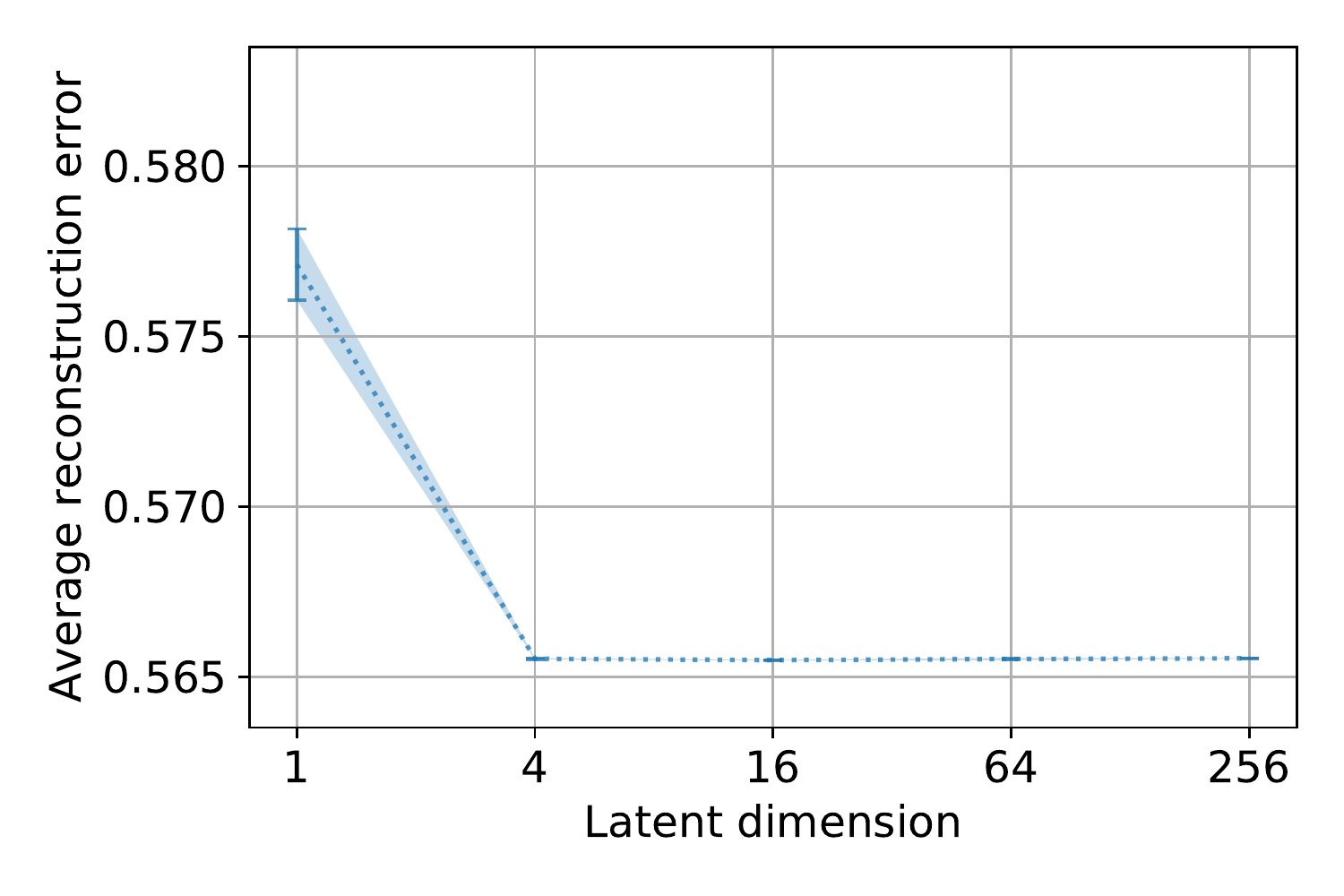}
        \caption{ObjectsRoom}
    \end{subfigure}
    \caption{The average reconstruction error on the validation set is fairly constant for sufficiently large latent dimensions and only degrades when a single latent is used.}
    \label{fig:reconerr_asymmetric}
\end{figure*}

\begin{figure}[h!]
    \centering
    \includegraphics[width=0.6\textwidth]{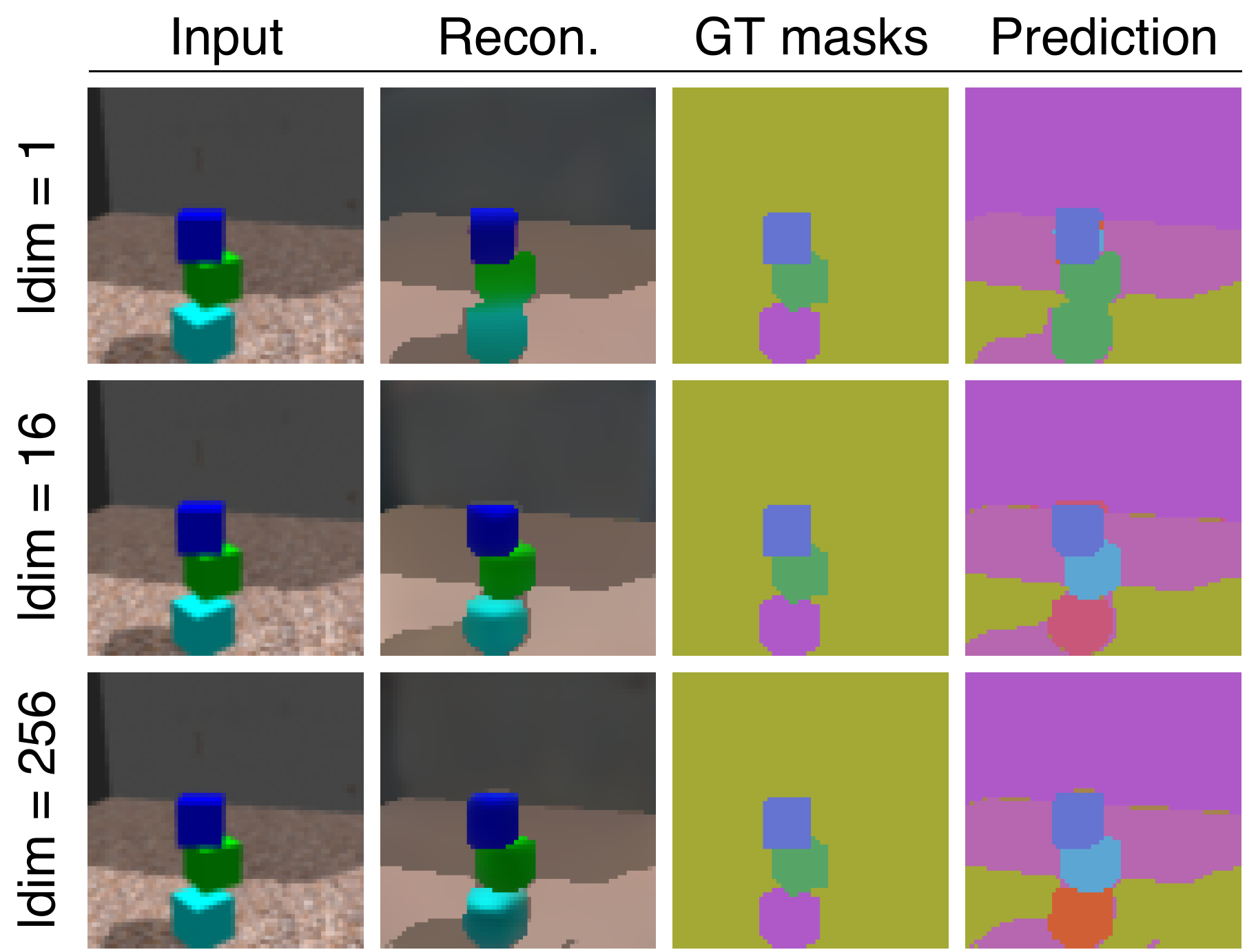}
    \caption{Qualitative segmentation performance with the original, asymmetric architecture on ShapeStacks for three different latent dimensions. Reconstruction and segmentation deteriorate slightly when only a single latent dimension is used.}
    \label{fig:qualitative_asymmetric}
\end{figure}

This architecture corresponds to the original \gls{GENESIS} formulation.
It is used for the results in Fig. \ref{fig:quantitative_asymmetric}, Fig. \ref{fig:reconerr_asymmetric}, and Fig. \ref{fig:qualitative_asymmetric} as well as the results for the corresponding vanilla \textsc{sbd-vae} in Tab. \ref{tab:bd}.
The latter is equivalent to training asymmetric \gls{GENESIS} with only a single component.

\begin{table}[h!]
    \centering
    \caption{Original \gls{VAE} encoder}
    \label{tab:bd_encoder}
    \begin{tabularx}{0.9\textwidth}{c c c c c c c }
        \toprule
        Input spatial dim & Input channels & Output channels & Kernel size & Stride & Layer type & Activation \\
        \midrule
        64 & 3 + 1 & 32   & 3$\times$3 & 2 & Conv & ELU \\
        32 & 32    & 32   & 3$\times$3 & 2 & Conv & ELU \\
        16 & 32    & 64   & 3$\times$3 & 2 & Conv & ELU \\
        8  & 64    & 64   & 3$\times$3 & 2 & Conv & ELU \\
        -  & 4$\times$4$\times$64 & 256  & -          & - & FC   & ELU \\
        -  & 256   & 2$\times$latent dim & - & - & FC  & - \\
        \bottomrule
    \end{tabularx}
\end{table}

\begin{table}[h!]
    \centering
    \caption{Original \gls{VAE} decoder}
    \label{tab:bd_decoder}
    \begin{tabularx}{0.9\textwidth}{c c c c c c c }
        \toprule
        Input spatial dim & Input channels & Output channels & Kernel size & Stride & Layer type & Activation \\
        \midrule
        - & latent dim & latent dim + 2 & - & - & Broadcast & - \\
        72 & latent dim + 2 & 32 & 3$\times$3 & 1 & Conv & ELU \\
        70 & 32 & 32 & 3$\times$3 & 1 & Conv & ELU \\
        68 & 32 & 32 & 3$\times$3 & 1 & Conv & ELU \\
        66 & 32 & 32 & 3$\times$3 & 1 & Conv & ELU \\
        64 & 32 & 3  & 1$\times$1 & 1 & Conv & - \\
        \bottomrule
    \end{tabularx}
\end{table}

\clearpage

\subsection{Symmetric \gls{GENESIS} and \textsc{dc-vae}}

\begin{figure*}[h!]
    \centering
    \begin{subfigure}{.33\textwidth}
        \includegraphics[trim=0 0 0 0, clip, width=\textwidth]{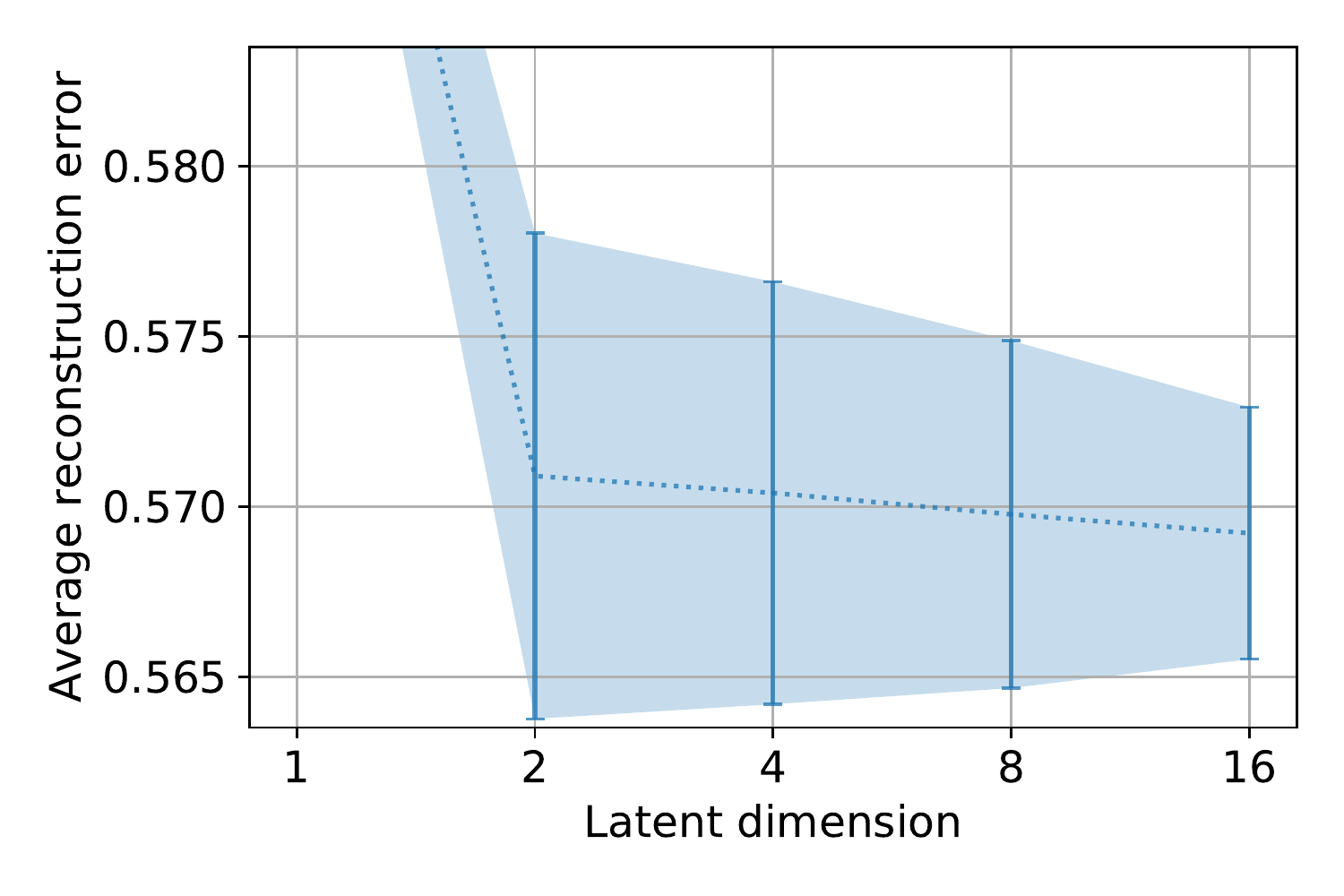}
        \caption{Multi-dSprites}
    \end{subfigure}
        \begin{subfigure}{.33\textwidth}
        \includegraphics[trim=0 0 0 0, clip, width=\textwidth]{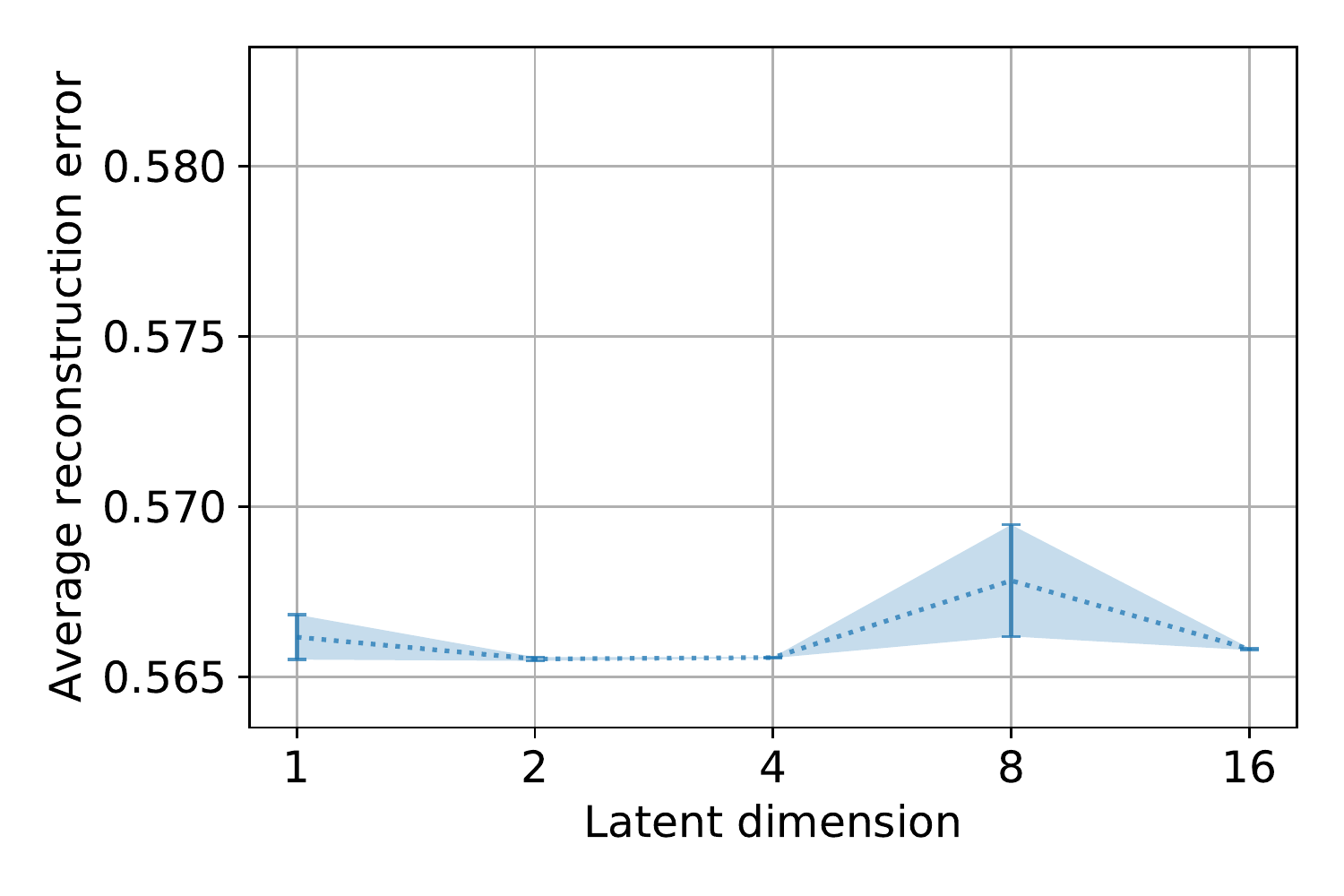}
        \caption{ShapeStacks}
    \end{subfigure}
    \begin{subfigure}{.33\textwidth}
        \includegraphics[trim=0 0 0 0, clip, width=\textwidth]{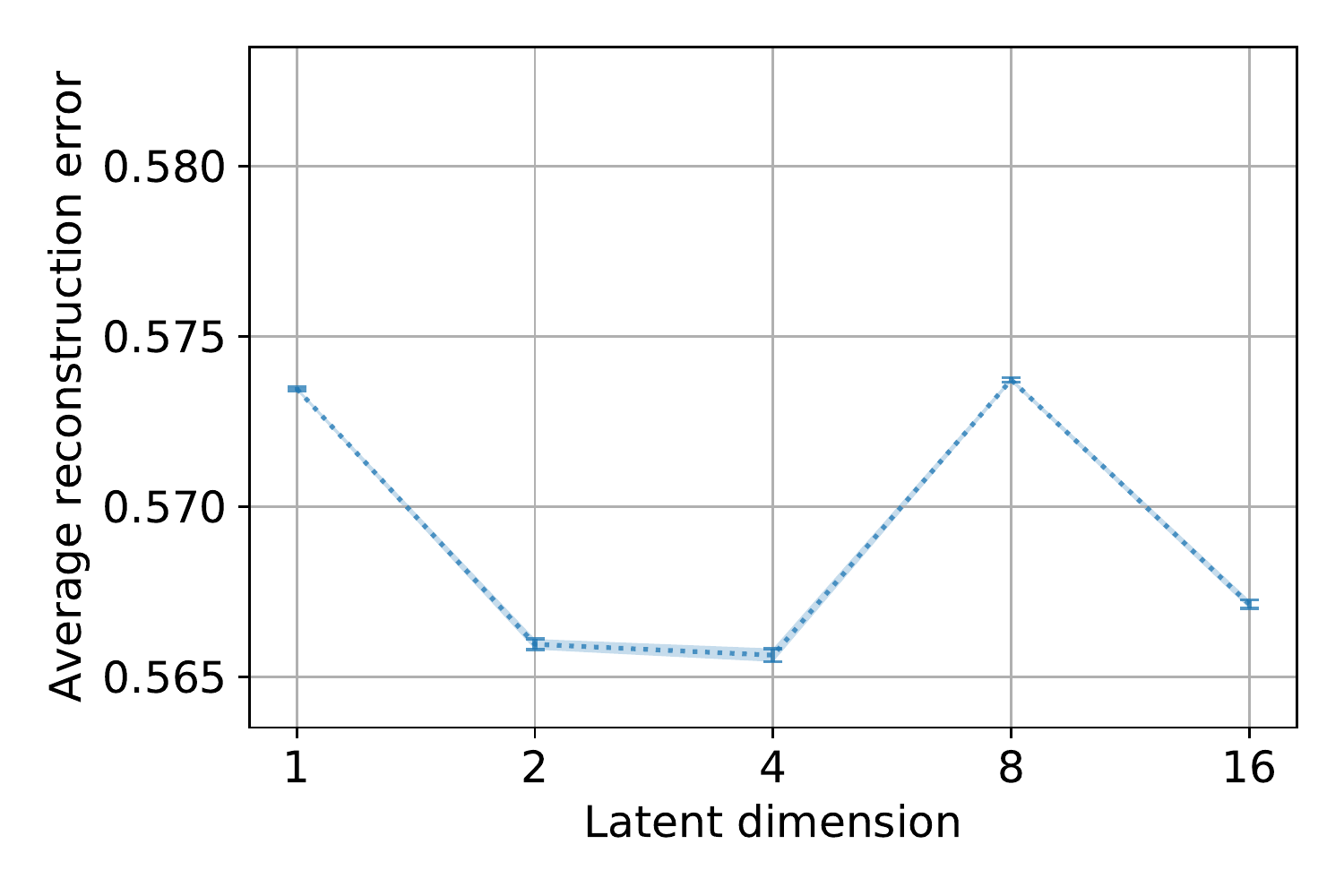}
        \caption{ObjectsRoom}
    \end{subfigure}
    \caption{Average reconstruction error on the validation set is less stable than for the asymmetric architecture, but also degrades when only a single latent is used.}
    \label{fig:reconerr_symmetric}
\end{figure*}

\begin{figure}[h!]
    \centering
    \includegraphics[width=0.6\textwidth]{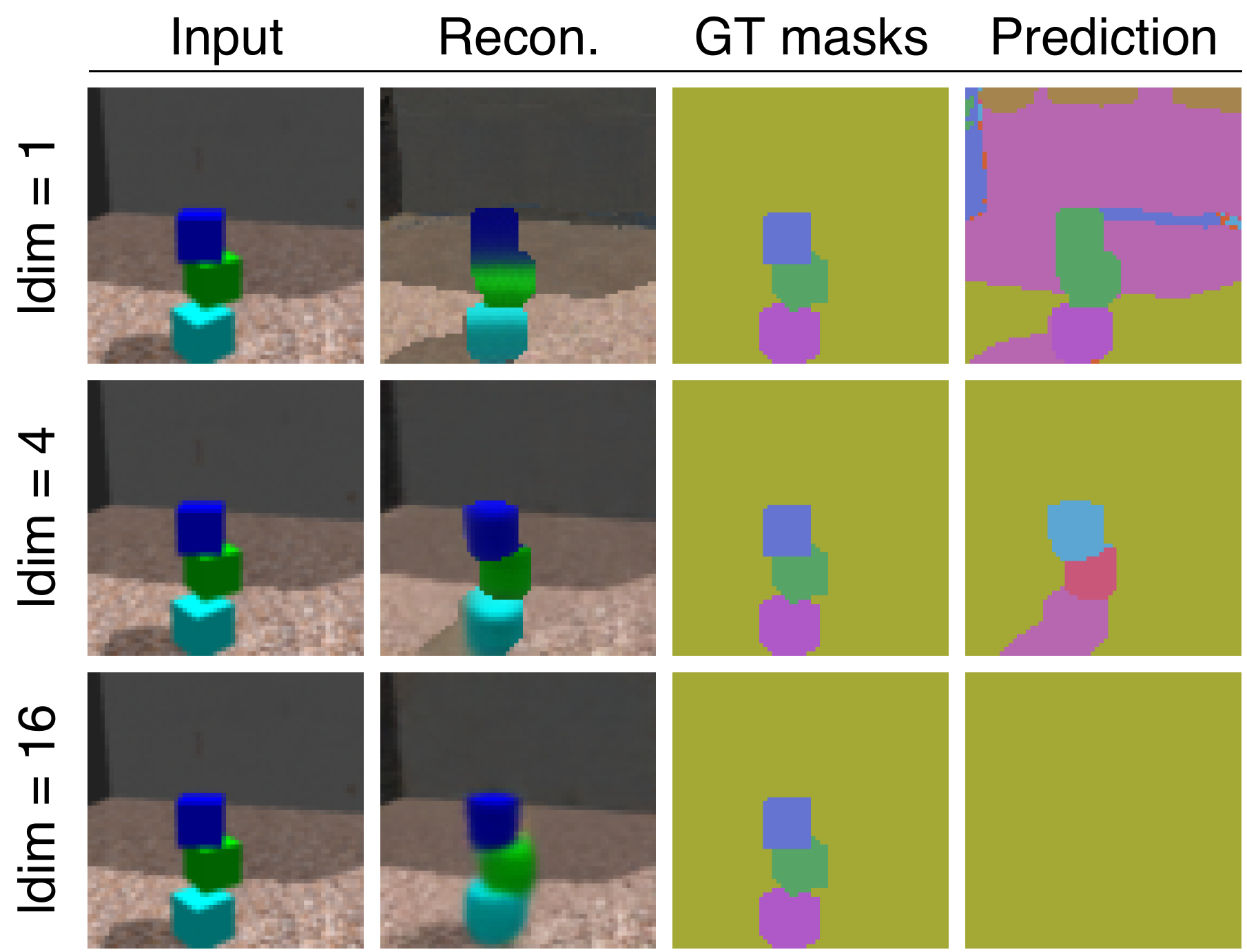}
    \caption{Qualitative segmentation performance with the modified, symmetric architecture on ShapeStacks for three different latent dimensions. Reconstruction and segmentation deteriorate when the latent dimension is too small. When the latent dimension is too large, segmentation collapses to a single component.}
    \label{fig:qualitative_symmetric}
\end{figure}

This architecture corresponds to the modified \gls{GENESIS} formulation.
It is used for the results in Fig. \ref{fig:quantitative_symmetric}, Fig. \ref{fig:reconerr_symmetric}, and Fig. \ref{fig:qualitative_symmetric} as well as the results for the corresponding vanilla \textsc{dc-vae} in Tab. \ref{tab:dc}.
The latter is equivalent to training symmetric \gls{GENESIS} with only a single component.
It encoder and decoder are very similar to the corresponding mask \gls{VAE} modules in \gls{GENESIS}.
Note that the GLU non-linearities half the number of output channels.

\begin{table}[h!]
    \centering
    \caption{Modified, higher capacity \gls{VAE} encoder}
    \label{tab:dc_encoder}
    \begin{tabularx}{0.9\textwidth}{c c c c c c c}
        \toprule
        Input spatial dim & Input channels & Output channels & Kernel size & Stride & Layer type & Activation \\
        \midrule
        64 & 3 + 1  & 64  & 5$\times$5 & 1 & Conv & BN-GLU \\
        64 & 32     & 32  & 5$\times$5 & 2 & Conv & BN-GLU \\
        32 & 32     & 128 & 5$\times$5 & 1 & Conv & BN-GLU \\
        32 & 64     & 128 & 5$\times$5 & 2 & Conv & BN-GLU \\
        16 & 64     & 128 & 5$\times$5 & 1 & Conv & BN-GLU \\
        -  & 16$\times$16$\times$64 & 4$\times$latent dim & - & - & FC & GLU \\
        \bottomrule
    \end{tabularx}
\end{table}

\begin{table}[h!]
    \centering
    \caption{Modified, higher capacity \gls{VAE} decoder}
    \label{tab:dc_decoder}
    \begin{tabularx}{0.9\textwidth}{c c c c c c c}
        \toprule
        Input spatial dim & Input channels & Output channels & Kernel size & Stride & Layer type & Activation \\
        \midrule
        - & latent dim & 16$\times$16$\times$128 & - & - & FC & GLU \\
        16 & 64 & 128 & 5$\times$5 & 1 & Deconv & BN-GLU \\
        16 & 64 & 64 & 5$\times$5 & 2 & Deconv & BN-GLU \\
        32 & 32 & 64 & 5$\times$5 & 1 & Deconv & BN-GLU \\
        32 & 32 & 64 & 5$\times$5 & 2 & Deconv & BN-GLU \\
        64 & 32 & 64 & 5$\times$5 & 1 & Deconv & BN-GLU \\
        64 & 32 & 3  & 1$\times$1 & 1 & Conv & - \\
        \bottomrule
    \end{tabularx}
\end{table}

\clearpage

\section{Reconstruction error}

\subsection{Asymmetric architecture}

\subsection{Symmetric architecture}

\clearpage

\section{Qualitative results}

\subsection{Asymmetric architecture}

\subsection{Symmetric architecture}

\end{document}